\documentclass[letterpaper, 10 pt, conference]{ieeeconf}
\IEEEoverridecommandlockouts                              
\usepackage{amsmath,amssymb}
\usepackage{comment}
\usepackage{url}

\usepackage{multirow}
\usepackage{graphicx}
\usepackage{epsfig}
\usepackage{epic,eepic}
\usepackage{times}
\usepackage{mathptmx}
\usepackage{cite}
\usepackage{color}

\usepackage{times}

\definecolor{red}{rgb}{1,0,0}
\definecolor{green}{rgb}{0,1,0}
\definecolor{blue}{rgb}{0,0,1}
\definecolor{violet}{rgb}{1,0,1}
\definecolor{cyan}{cmyk}{1,0,0,0}
\definecolor{magenta}{cmyk}{0,1,0,0}
\definecolor{yellow}{cmyk}{0,0,1,0}

\definecolor{white}{rgb}{1,1,1}

\usepackage{arydshln}

\newcommand{\CO}[1]{}

\newcommand{\CommentOut}[1]{}

 \newcommand{\editage}[1]{}

\twocolumn

\begin{document}

\title{\LARGE\bf%
Pole-centric Descriptors for Robust Robot Localization: Evaluation under Pole-at-Distance (PaD) Observations using the Small Pole Landmark (SPL) Dataset
}

\author{Wuhao Xie \and Kanji Tanaka
\thanks{Our work has been supported in part by JSPS KAKENHI Grant-in-Aid for Scientific Research (C) 20K12008 and 23K11270.}
\thanks{%
W. Xie and K. Tanaka are with Fundamental Engineering for Knowledge-Based Society, Graduate School of Engineering, University of Fukui, Japan. 
{\tt\small{\{mf250169, tnkknj\}@g.u-fukui.ac.jp}}
}}

\maketitle

\title{\LARGE \bf
Robustness Evaluation of Pole-centric Descriptors under Sparse and Long-range Observations using the SPL Dataset
}

\author{Author Name$^{1}$
\thanks{$^{1}$Author is with the Faculty of ...}
}

\maketitle
\thispagestyle{empty}
\pagestyle{empty}

\begin{abstract}
While pole-like structures are widely recognized as stable geometric anchors for long-term robot localization, their identification reliability degrades significantly under Pole-at-Distance (Pad) observations typical of large-scale urban environments. This paper shifts the focus from descriptor design to a systematic investigation of descriptor robustness. Our primary contribution is the establishment of a specialized evaluation framework centered on the Small Pole Landmark (SPL) dataset. This dataset is constructed via an automated tracking-based association pipeline that captures multi-view, multi-distance observations of the same physical landmarks without manual annotation. Using this framework, we present a comparative analysis of Contrastive Learning (CL) and Supervised Learning (SL) paradigms. Our findings reveal that CL induces a more robust feature space for sparse geometry, achieving superior retrieval performance particularly in the 5--10m range. This work provides an empirical foundation and a scalable methodology for evaluating landmark distinctiveness in challenging real-world scenarios.
\end{abstract}

\section{INTRODUCTION}

The reliability of landmark-based SLAM and map maintenance hinges on the ability to consistently identify features across different sessions and viewpoints \cite{Meyer2019SurveySLAM}. Among various geometric features, poles are highly advantageous due to their long-term structural stability and ease of detection. However, they suffer from low distinctiveness, leading to perceptual aliasing in repetitive urban environments.

To address this, recent research has introduced pole-centric descriptors that encode the local 3D context around a pole. While these representations show promise, existing studies have primarily focused on "proposal-stage" evaluations under relatively dense point cloud conditions. There is a critical knowledge gap regarding how these descriptors behave under the "small-scale" observations prevalent in long-range sensing, where point density is extremely low and geometry is fragmented \cite{9564759}.

In this paper, we argue that the advancement of robust localization requires not just new descriptors, but a rigorous evaluation of existing learning paradigms under these edge-case conditions. Consequently, the core of this work is the construction of the Small Pole Landmark (SPL) dataset and a comparative analysis of learning objectives. We specifically investigate whether self-supervised Contrastive Learning (CL) \cite{He2020MoCo} offers better generalization to distance-induced sparsity compared to standard Supervised Learning (SL).

The key contributions of this paper, distinct from prior methodological proposals, are:
\begin{itemize}
    \item \textbf{Construction of the SPL Dataset}: An automated pipeline that leverages temporal tracking to generate thousands of matched landmark pairs under varying degrees of sparsity, providing a benchmark for "small-scale" landmark identification.
    \item \textbf{Empirical Analysis of CL vs. SL}: A quantitative demonstration that CL naturally learns viewpoint-invariant features that are more resilient to the structural degradation caused by long-range sensing.
    \item \textbf{Systematic Robustness Breakdown}: An analysis of the trade-off between observation distance and descriptor reliability, identifying the effective operational range for pole-centric localization in urban environments.
\end{itemize}

\section{System Context and Rationale}

The descriptor evaluated in this study is designed as a core component of a broader framework for long-term autonomous navigation. This section outlines the underlying logic of using poles as anchors and the integrated system architecture.

\subsection{Poles as Geometric Anchors and Logical Rationale}
The proposed approach leverages poles as high-precision "anchors" to identify unique "signatures" from the surrounding 3D structure \cite{xie2025poleimageselfsupervisedpoleanchoreddescriptor}. Unlike general landmarks where detection and identification are often intertwined, pole-like structures allow these tasks to be decoupled. Their simple geometry and verticality make "detection" alone extremely stable and easy, providing a powerful mechanism for trackability. This ease of data acquisition enables the automatic, large-scale collection of diverse multi-view observations (positive pairs) belonging to the same physical landmark, providing a strong logical rationale for applying Contrastive Learning (CL) to this problem.

\subsection{System Architecture and Dual-purpose Utility}
The Pole-Image descriptor is integrated into the \textbf{APD-PSL-SSL} architecture \cite{xie2025poleimageselfsupervisedpoleanchoreddescriptor}, a closed-loop framework consisting of: (i) \textbf{Active Policy Determination (APD)} for long-range detection and approach; (ii) \textbf{Passive Self-Localization (PSL)} for local landmark identification; and (iii) \textbf{Self-Supervised Learning (SSL)} for continuous encoder refinement via pole tracking. 

While the full architectural details are provided in \cite{xie2025poleimageselfsupervisedpoleanchoreddescriptor}, the scope of this paper is strictly the \textbf{robustness evaluation of the PSL component} under the sparse, long-range conditions that typically degrade geometric signatures. This evaluation is critical because the descriptor serves a dual purpose: its distinctiveness enables robust self-localization against perceptual aliasing, while its high-precision geometry facilitates sensitive environmental change detection for long-term map maintenance.

\section{SPL DATASET CONSTRUCTION AND ANALYSIS}

The primary contribution of this work is the methodology for constructing a robustness-centric dataset without manual effort.

\subsection{NCLT Dataset}

This study employs the North Campus Long-Term (NCLT) Dataset \cite{carlevaris2016university}, collected using a Segway-based robotic platform at the University of Michigan's North Campus, as the foundational benchmark for evaluation. This dataset was acquired using a suite of sensors, including a Velodyne HDL-32E LiDAR, across diverse seasons, weather conditions, and traversal routes. It is widely recognized as a standard benchmark for validating the long-term environmental robustness of autonomous mobile robot self-localization and mapping. In this paper, we construct the Small Pole Landmark (SPL) Dataset from the NCLT data, specifically targeting small-scale cylindrical structures such as utility poles and signposts located at a distance. More specifically, we aggregate sparse and partially observed point cloud data from multiple viewpoints and time instances for identical poles, utilizing a tracking-based association across consecutive LiDAR frames. 
Specific sessions were selected for training, while an entirely different session---collected under different seasonal or atmospheric conditions---was designated for testing.

\subsection{Tracking-based Automated Labeling}

We utilize a LiDAR-based tracking framework to associate pole detections across time. For pole detection, we employ Polex \cite{Schaefer2022Polex}, which combines geometric heuristics with a segmentation pipeline to extract vertical pole-like structures. Its accuracy and efficiency in urban LiDAR scans make it well-suited for this task. By applying an adaptive spatial gating strategy, we can maintain the identity of a landmark from its first long-range detection through to a close-range encounter. This allows us to automatically label distant, sparse observations (queries) with their corresponding dense, close-range counterparts (database entries).

\subsection{Dataset Characteristics}
The SPL dataset is specifically filtered to include "Small Pole Landmarks"---those observed at distances where the total point count is below a threshold (10). This results in a challenging set of query-positive pairs that directly represent the noise and sparsity of real-world urban deployments.

\subsection{Baseline Descriptor: Pole-Image}
We adopt the Pole-Image representation \cite{xie2025poleimageselfsupervisedpoleanchoreddescriptor}, which transforms the local geometry around a pole into a 2D signature. First, we identify pole centroids using Polex \cite{Schaefer2022Polex} as reference points. To ensure rotational invariance, the surrounding point cloud is transformed into a pole-centric cylindrical coordinate system $(r, \theta, z)$. These points are then projected onto an $80 \times 360$ 2D grayscale image. This "Iris-style" projection converts the unstructured 3D point cloud into a canonical 2D format suitable for standard convolutional neural networks.

\subsection{Learning Strategy and Network Architecture}
For the feature extraction stage, we employ a lightweight \textbf{ResNet-based encoder} to map the generated Pole-Images into a discriminative 128-D embedding space. To investigate the impact of the learning objective on descriptor robustness, we train two identical architectures using different paradigms:
\begin{enumerate}
    \item \textbf{Supervised Learning (SL)}: Each pole track ID is treated as a distinct class for cross-entropy optimization.
    \item \textbf{Contrastive Learning (CL)}: Pairs from the same track are pulled together using the InfoNCE loss ($\tau = 0.07$), while pairs from different tracks are pushed apart.
\end{enumerate}

For the Contrastive Learning (CL) regime, we adopt the NT-Xent loss (also known as InfoNCE) to effectively learn instance-level invariance. The temperature parameter is set to $\tau = 0.07$, which serves to scale the cosine similarities and control the hardness of negative samples in the mini-batch. The models are optimized using the Adam optimizer with a learning rate of $10^{-3}$ over 30 epochs to ensure stable convergence.

\section{EXPERIMENTAL RESULTS AND INSIGHTS}

\subsection{Evaluation Setup}

In this study, we evaluate the robustness of the pole-centric descriptor under Pole-at-Distance (PaD) observations using the SPL dataset. For the retrieval evaluation, we utilized two complete single-season traversals from the NCLT dataset: the "2012-05-26" session for training and the "2012-08-04" session for testing. A total of 359 unique pole landmarks were identified as common to both sessions. Although each physical landmark is observed multiple times within the training session due to temporal tracking, we constructed the reference database by randomly sampling a single instance from each corresponding track. This resulted in a database of 359 distinct entries, serving as the ground truth for our identification task. This setup facilitates a comprehensive assessment of landmark recognition across the entire environment under varying seasonal conditions.

\subsection{Overall Retrieval Performance}

The retrieval task requires finding the correct landmark in a database of 3D signatures from a previous session. For this evaluation, we utilized a complete single-season traversal from the NCLT dataset as the test session, which contains a total of 359 unique pole landmarks. As shown in Table \ref{tab:spl_overall_recall}, CL significantly outperforms SL.

\begin{table}[t]
\centering
\caption{Overall retrieval performance on the SPL dataset.}
\label{tab:spl_overall_recall}
\begin{tabular}{lcccc}
\hline
Method & \#Query & R@1 & R@5 & R@10 \\
\hline
Contrastive Learning (CL) & 359 & \textbf{22.01} & \textbf{47.91} & \textbf{61.00} \\
Supervised Learning (SL)  & 359 & 17.55 & 44.85 & 59.61 \\
\hline
\end{tabular}
\end{table}

\subsection{Distance-induced Robustness Analysis}
A key finding of our evaluation is the breakdown of performance by observation range (Table \ref{tab:spl_range_binned_recall}). 

\begin{table}[t]
\centering
\caption{Retrieval performance grouped by observation range.}
\label{tab:spl_range_binned_recall}
\begin{tabular}{llcccc}
\hline
Method & Range & \#Query & R@1 & R@5 & R@10 \\
\hline
CL & $\leq 5$m & 40 & \textbf{37.50} & \textbf{75.00} & \textbf{85.00} \\
   & $(5,10]$m & 153 & \textbf{20.92} & \textbf{46.41} & \textbf{62.09} \\
\hline
SL & $\leq 5$m & 40 & 20.00 & 60.00 & 75.00 \\
   & $(5,10]$m & 153 & 15.03 & 40.52 & 56.86 \\
\hline
\end{tabular}
\end{table}

\subsection{Discussion: Why CL Excels}
The data suggests that CL is more effective at learning the "topology" of the local environment rather than the specific "appearance" of a class. In the 5--10m range, where viewpoint changes are significant but some structure remains, CL's instance-discrimination objective encourages the network to ignore the sampling noise inherent in sparse LiDAR scans, a property that SL lacks as it focuses on inter-class boundaries.


\section{RELATED WORK}

\subsection{Evaluation of 3D Descriptors}
Traditional evaluation of 3D descriptors (e.g., FPFH \cite{Rusu2009FPFH}, SHOT \cite{salti2014shot}) often relies on dense mesh or high-resolution LiDAR data. However, in mobile robotics, the challenge lies in sparsity. Recent benchmarks have begun to address cross-session localization \cite{9564759}, but a specific focus on the robustness of pole-centric anchors under long-range degradation remains sparse.

\subsection{Contrastive vs. Supervised Learning}
The choice of learning objective is critical for feature robustness. While Supervised Learning (SL) optimizes for fixed class separation, Contrastive Learning (CL) \cite{He2020MoCo, chen2020simple} focuses on instance-level invariance. This work evaluates these paradigms specifically for the task of landmark identification in sparse 3D environments, a domain where the benefits of CL have not been fully quantified against SL.

\section{CONCLUSION AND FUTURE WORK}

In this paper, we moved beyond the proposal of pole descriptors to a rigorous evaluation of their robustness. By constructing the SPL dataset through an automated tracking-based pipeline, we have provided a framework for evaluating landmark identification in realistic, sparse environments. Our comparative analysis identifies Contrastive Learning as a superior paradigm for training descriptors resilient to long-range degradation.

Future work will build on this evaluation framework by:
\begin{itemize}
    \item \textbf{Expanding Environmental Diversity}: Incorporating suburban and park environments where pole landmarks are even more sparse and ambiguous.
    \item \textbf{Temporal Robustness Evaluation}: Extending the SPL dataset to cover multi-seasonal data to evaluate the impact of appearance changes (e.g., foliage) on pole-centric signatures.
    \item \textbf{Generalizing to Non-Pole Anchors}: Applying this evaluation methodology to other stable geometric anchors like building corners or planar segments.
\end{itemize}

\bibliography{reference} 
\bibliographystyle{unsrt} 

\end{document}